# Robot Location Estimation in the Situation Calculus

**Vaishak Belle** and **Hector J. Levesque**
Dept. of Computer Science
University of Toronto
Toronto, Ontario M5S 3H5, Canada
{vaishak, hector}@cs.toronto.edu

## Abstract

Location estimation is a fundamental sensing task in robotic applications, where the world is uncertain, and sensors and effectors are noisy. Most systems make various assumptions about the dependencies between state variables, and especially about how these dependencies change as a result of actions. Building on a general framework by Bacchus, Halpern and Levesque for reasoning about degrees of belief in the situation calculus, and a recent extension to it for continuous domains, in this paper we illustrate location estimation in the presence of a rich theory of actions using an example. We also show that while actions might affect prior distributions in nonstandard ways, suitable posterior beliefs are nonetheless entailed as a side-effect of the overall specification.

## Introduction

Location estimation is a fundamental sensing task in robotic applications (Thrun, Burgard, and Fox 2005), where the world is uncertain, and sensors and effectors are noisy. Agents operating under these conditions grapple with at least two sorts of reasoning problems. First, because the world is *dynamic*, actions perpetually change the properties of the state. Second, because little in the world is definite, the agent has to modify its beliefs based on the actions performed and the results returned by its sensors.

To see a simple example, imagine a robot located in a 2-dimensional grid, at a certain distance $h$ to the right of a wall, as in Figure 1. The robot might initially believe that $h$ is drawn from a uniform distribution on [2, 12]. Among the robot's many capabilities, we might imagine the ability of moving left. A leftwards motion of 1 unit would shift the uniform distribution on $h$ to [1, 11], but a leftward motion of 4 units would change the distribution more radically. The point $h = 0$ would now obtain a weight of .2, while $h \in (0, 8]$ would retain their densities. This mixed distribution would then be preserved by a subsequent rightward motion. Likewise, we might imagine the robot to be equipped with two onboard sensors: a sonar unit aimed at the wall estimating $h$, and a GPS (global positioning system) device sensing both $h$ and the robot's vertical position. Each of these might be characterized by Gaussian error models, and the effect of a reading from any sensor would revise the distribution on $h$ from uniform to an appropriate Gaussian. The robot is now left with the difficult task of adjusting its beliefs as it moves and obtains competing (perhaps conflicting) measurements from individual sensors.

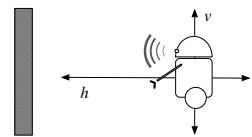

Figure 1: Robot operating in a 2-dimensional world.

Probabilistic formalisms such as Kalman filtering (Dean and Wellman 1991; Fox et al. 2003), and more generally, Dynamic Bayesian Networks (Dean and Kanazawa 1989; Pearl 1988), address sensor fusion. But while belief update of known priors wrt Gaussian and other continuous error models is treated appropriately, very little is said about how actions might change values of certain state variables while not affecting others. These formalisms also assume a full specification of the dependencies between variables, making it difficult to deal with other forms of incomplete knowledge, and strict uncertainty in particular. Therefore occurrences such as shifting densities (such as in the example above) and shifting dependencies (assume lateral motion depends on the ground's slipperiness) are hard to address in a general way.

Perhaps the most general formalism for dealing with probabilistic belief in formulas, and how that should evolve in the presence of noisy acting and sensing, is a logical account by Bacchus, Halpern and Levesque (BHL) (1999). In the BHL approach, besides quantifiers and other logical connectives, one has the provision for specifying the *degrees of belief* in formulas in the initial state. This specification may be compatible with one or very many initial distributions and sets of independence assumptions. All the properties of belief will then follow at a corresponding level of specificity.

Subjective uncertainty is captured in the BHL scheme using a possible-world model of belief (Kripke 1963; Hintikka 1962; Fagin et al. 1995). Intuitively, the degree of belief in $\phi$ is defined as a normalized sum over the possible worlds where $\phi$ is true of some nonnegative *weights* associated with those worlds. To reason about belief change, the BHL model is then embedded in a rich theory of action and sensing provided by the situation calculus (McCarthy and Hayes 1969;

Reiter 2001; Scherl and Levesque 2003). The BHL account provides axioms in the situation calculus regarding how the weight associated with a possible world changes as the result of acting and sensing. The properties of belief and belief change then emerge as a direct logical consequence of the initial specifications and these changes in weights.

However, in contrast to the earlier mentioned Bayesian formalisms, one of the limitations of the BHL approach is that it is restricted to fluents whose values are drawn from discrete countable domains. One could say, for example, that $h \in \{2, 3, \ldots, 11\}$ is given an equal weight of .1, but stipulating a continuous uniform distribution and Gaussian sensor error models (and not discrete approximations thereof) is quite beyond the BHL approach. In (Belle and Levesque 2013), we show how with minimal additional assumptions this serious limitation of BHL can be lifted.

In this paper, we illustrate how the (generalized) BHL scheme is utilized in location estimation using our example consisting of a robot's position in *XY*-plane, a sonar and a GPS device. Our example supposes that the robot is capable of deterministic physical actions, while the sensors are characterized by continuous error models. We stipulate that the GPS device operates problematically when the robot approaches the wall, perhaps due to signal obstructions, in which case readings are subject to systematic bias. Thus, the domain formalization illustrates belief change wrt shifting densities as logical properties of actions, competing sensors, and situation-specific bias, among others. Since no assumptions need to be made in general regarding the kind of distributions that initial state variables are drawn from, nor about dependencies between state variables, this work illustrates how beliefs about the robot's location would change after acting and sensing in complex uncertain domains.

The paper is structured as follows. In the next section, we briefly review formal preliminaries, such as the situation calculus and the BHL scheme, as well as the essentials of its generalization to continuous domains. We then model the robot domain and illustrate properties about belief change. In the final sections, we discuss related and future work.

## Preliminaries

The language $\mathcal{L}$ of the situation calculus (McCarthy and Hayes 1969) is a many-sorted dialect of predicate calculus, with sorts for *actions*, *situations* and *objects* (for everything else, and includes the set of reals $\mathbb{R}$ as a subsort). A situation represents a world history as a sequence of actions. A set of initial situations correspond to the ways the world might be initially. Successor situations are the result of doing actions, where the term $do(a, s)$ denotes the unique situation obtained on doing $a$ in $s$. The term $do(\alpha, s)$, where $\alpha$ is the sequence $[a_1, \ldots, a_n]$ abbreviates $do(a_n, do(\ldots, do(a_1, s) \ldots))$. Initial situations are defined as those without a predecessor:

$$Init(s) \doteq \neg \exists a, s'. \, s = do(a, s').$$

We let the constant $S_0$ denote the actual initial situation, and we use the variable $\iota$ to range over initial situations only.

In general, the situations can be structured into a set of trees, where the root of each tree is an initial situation and the edges are actions. In dynamical domains, we want the values of predicate and functions to vary from situation to situation. For this purpose, $\mathcal{L}$ includes *fluents* whose last argument is always a situation. Here we assume without loss of generality that all fluents are functional.

**Basic action theory**   Following (Reiter 2001), we model dynamic domains in $\mathcal{L}$ by means of a *basic action theory* $\mathcal{D}$, which consists of [1]

1. axioms $\mathcal{D}_0$ that describe what is true in the initial states, including $S_0$;

2. precondition axioms that describe the conditions under which actions are executable;

3. successor state axioms that describe the changes to fluents on executing actions;

4. domain-independent *foundational* axioms, the details of which need not concern us here. See (Reiter 2001).

An agent reasons about actions by means of the entailments of $\mathcal{D}$, for which standard Tarskian models suffice. We assume henceforth that models *also* assign the usual interpretations to $=, <, >, 0, 1, +, \times, /, -, e, \pi$ and $x^y$ (exponentials).[2]

**Likelihood and degree of belief**   The BHL model of belief builds on a treatment of knowledge by Scherl and Levesque (2003). Here we present a simpler variant based on just two distinguished binary fluents $l$ and $p$.

The term $l(a, s)$ is intended to denote the likelihood of action $a$ in situation $s$. For example, suppose $sonar(z)$ is the action of reading the value $z$ from a sensor that measures the distance to the wall, $h$.[3] We might assume that this action is characterized by a Gaussian error model:[4]

$$l(sonar(z), s) = u \equiv$$
$$(z \geq 0 \wedge u = \mathcal{N}(z - h(s); \mu, \sigma^2)) \vee (z < 0 \wedge u = 0)$$

which stipulates that the difference between a nonnegative reading of $z$ and the true value $h$ is normally distributed with a variance of $\sigma^2$ and mean of $\mu$. In general, the action theory $\mathcal{D}$ is assumed to contain for each action type $A$ an additional *action likelihood axiom* of the form

$$l(A(\vec{x}), s) = u \equiv \phi_A(\vec{x}, u, s)$$

where $\phi_A$ is a formula that characterizes the conditions under which action $A(\vec{x})$ has likelihood $u$ in $s$. (Actions that have no sensing aspect should be given a likelihood of 1.)

Next, the $p$ fluent determines a probability distribution on situations. The term $p(s', s)$ denotes the relative *weight* accorded to situation $s'$ when the agent happens to be in situation $s$. The properties of $p$ in initial states, which vary from domain to domain, are specified by axioms as part of $\mathcal{D}_0$. The following nonnegative constraint is also included in $\mathcal{D}_0$:

$$\forall \iota, s.\ p(s, \iota) \geq 0 \wedge (p(s, \iota) > 0 \supset Init(s)) \quad (P1)$$

---

[1] As usual, free variables in any of these axioms should be understood as universally quantified from the outside.

[2] Alternatively, one could specify axioms for characterizing the field of real numbers in $\mathcal{D}$. Whether or not reals with exponentiation is *first-order* axiomatizable remains a major open question.

[3] Naturally, we assume that the value $z$ being read is not under the agent's control. See BHL for a precise rendering of this nondeterminism in terms of GOLOG operators (Reiter 2001).

[4] Note that $\mathcal{N}$ is a continuous distribution involving $\pi$, $e$, exponentiation, and so on. Therefore, BHL always consider discrete probability distributions that *approximate* the continuous ones.

While this is a stipulation about initial states $\iota$ only, BHL provide a successor state axiom for $p$, and show that with an appropriate action likelihood axiom, the nonnegative constraint then continues to hold everywhere:

$$\begin{aligned} p(s', do(a, s)) = u \equiv \\ \exists s''\ [s' = do(a, s'') \wedge Poss(a, s'') \wedge \\ u = p(s'', s) \times l(a, s'')] \\ \vee \neg \exists s''\ [s' = do(a, s'') \wedge Poss(a, s'') \wedge u = 0] \end{aligned} \quad (P2)$$

Now if $\phi$ is a formula with a single free variable of sort situation,[5] then the *degree of belief* in $\phi$ is simply defined as the following abbreviation:

$$Bel(\phi, s) \doteq \frac{1}{\gamma} \sum_{\{s':\phi[s']\}} p(s', s) \quad (B)$$

where $\gamma$, the normalization factor, is understood throughout as the same expression as the numerator but with $\phi$ replaced by *true*. For example, here $\gamma$ is $\sum_{s'} p(s', s)$. We do not have to insist that $s'$ and $s$ share histories since $p(s', s)$ will be 0 otherwise. BHL show how summations can be expressed using second-order logic, see the appendix. That is, neither *Bel*'s definition nor summations are special axioms of $\mathcal{D}$, but simply convenient abbreviations for logical terms. To summarize, in the BHL scheme, an action theory consists of:

1. $\mathcal{D}_0$ as before, but now also including (P1);
2. precondition axioms as before;
3. successor state axioms as before, but now also including one for $p$ viz. (P2);
4. foundational domain-independent axioms as before; and
5. action likelihood axioms.

**From sums to integrals** While the definition of belief in BHL has many desirable properties, it is defined in terms of a *summation* over situations, and therefore precludes fluents whose values range over the reals. The continuous analogue of (B) then requires *integrating* over some suitable space of values.

As it turns out, a suitable space can be found. First, some notation. We use a form of conditional *if-then-else* expressions, by taking some liberties with notation and the scope of variables as follows. We write $f = \text{IF}\ \exists x.\ \phi\ \text{THEN}\ t_1\ \text{ELSE}\ t_2$ to mean the logical formula

$$f = u \equiv \exists x.\ [\phi \wedge (u = t_1)] \vee [(u = t_2) \wedge \neg \exists x.\ \phi]$$

Now, assume that there are $n$ fluents $f_1, \ldots, f_n$ in $\mathcal{L}$, and that these take no arguments other than a situation.[6] Next, suppose that there is exactly one initial situation for any vector of fluent values (Levesque, Pirri, and Reiter 1998):

$$[\forall \vec{x} \exists \iota \bigwedge f_i(\iota) = x_i] \wedge [\forall \iota, \iota'. \bigwedge f_i(\iota) = f_i(\iota') \supset \iota = \iota'] \quad (*)$$

---

[5]The $\phi$ is usually written either with the situation variable suppressed or with a distinguished variable *now*. Either way, $\phi[t]$ is used to denote the formula with that variable replaced by $t$.

[6]Basically, if we were to assume that the arguments of all fluents, even $k$-ary ones, are taken from finite sets then this would allow us to enumerate the $n$ random variables of the domain (for some large $n$). Note that, from the point of view of situation calculus basic action theories, fluents are typically allowed to take *arguments* from any set, including infinite ones. In probabilistic terms, this would this would correspond to having a joint probability distribution over infinitely many, perhaps uncountably many, random variables. We know of no existing work of this sort, and we have as yet no good ideas about how to deal with it.

Under these assumptions, it can be shown that the summation over all situations in (B) can be recast as a summation over all possible initial values $x_1, \ldots, x_n$ for the fluents:

$$Bel(\phi, s) \doteq \frac{1}{\gamma} \sum_{\vec{x}} P(\vec{x}, \phi, s) \quad (B')$$

where $P(\vec{t}, \phi, s)$ is the (unnormalized) weight accorded to the *successor* of an initial world where $f_i$ equals $t_i$:

$$\begin{aligned} P(\vec{t}, \phi, do(\alpha, S_0)) \doteq \\ \quad \text{IF}\ \exists \iota.\ \bigwedge f_i(\iota) = t_i \wedge \phi[do(\alpha, \iota)] \\ \quad \quad \text{THEN}\ \ p(do(\alpha, \iota), do(\alpha, S_0)) \\ \quad \text{ELSE}\ \ 0 \end{aligned}$$

where $\alpha$ is an action sequence. In a nutshell, because every situation has an initial situation as an ancestor, and because there is a bijection between initial situations and possible fluent values, it is sufficient to sum over fluent values to obtain the belief even for non-initial situations. Note that unlike (B), this one expects the final situation term $do(\alpha, S_0)$ mentioning what actions and observations took place to be explicitly specified, but that is just what one expects when the agent reasons about its belief after doing things, and for the projection problem in particular (Reiter 2001).

The generalization to the continuous case then proceeds as follows. First, we observe that some (though possibly not all) fluents will be real-valued, and that $p(s', s)$ will now be a measure of *density* not weight. Similarly, the *P* term above now measures (unnormalized) density rather than weight.

Now suppose fluents are partitioned into two groups: the first $k$ take their values $x_1, \ldots, x_k$ from $\mathbb{R}$, while the rest take their values $y_{k+1}, \ldots, y_n$ from countable domains. Then the *degree of belief* in $\phi$ is an abbreviation for:

$$Bel(\phi, s) \doteq \frac{1}{\gamma} \int_{\vec{x}} \sum_{\vec{y}} P(\vec{x} \cdot \vec{y}, \phi, s)$$

The belief in $\phi$ is obtained by ranging over all possible fluent values, and integrating and summing the densities of situations where $\phi$ holds.[7] In the appendix, we show how integrals can be formulated using second-order quantification. That is, as before, *Bel*, *P*, integrals and sums are simply convenient abbreviations, and do not involve special axioms in $\mathcal{D}$. More precisely, the continuous extension to BHL has the same components from earlier, with a single revision:

1. $\mathcal{D}_0$ additionally includes $(*)$.

Note that likelihood axioms are specified as before, but we will no longer have to approximate Gaussian error models (or any other continuous models) as would BHL.

---

[7]We are assuming here that the density function is (Riemann) integrable. If it is not, belief is clearly not defined, nor should it be. Similarly, if the normalization factor is 0, which corresponds to the case of conditioning on an event that has 0 probability, belief should not be (and is not) defined.

# Location Estimation: An Example
## Action Theory

We build a basic action theory $\mathcal{D}$ for a robot in a 2-dimensional grid. We imagine two fluents $h$ and $v$ in addition to *Poss*, $l$ and $p$. The fluent $h$ gives the distance to the wall and $v$ gives the position of the robot along the vertical axis. We consider two physical actions *left*($z$) and *up*($z$), and two sensing actions *sonar*($x$) and *gps*($x, y$).

$\mathcal{D}_0$ includes the following domain-independent axioms: (∗) and (P1). Specific to the domain, imagine that $\mathcal{D}_0$ also includes the following for $p$:

$$p(\iota, S_0) = \begin{cases} .1 \times \mathcal{N}(v(\iota); 0, 16) & \text{if } 2 \leq h(\iota) \leq 12 \\ 0 & \text{otherwise} \end{cases}$$

This says that the value of $v$ is normally distributed about the horizontal axis with variance 16, and independently, that the value of $h$ is uniformly distributed between 2 and 12.[8] No other sentence is included in $\mathcal{D}_0$.

For simplicity, we assume that actions are always executable. Therefore, $\mathcal{D}$ will not contain any precondition axioms. $\mathcal{D}$'s successor state axioms are the following. There is a fixed one for $p$, which is (P2). For $h$, we have:

$$\begin{aligned} h(do(a, s)) = u \equiv \\ \neg \exists z(a = \mathit{left}(z)) \wedge u = h(s) \vee \\ \exists z(a = \mathit{left}(z) \wedge u = \max(0, h(s) - z)). \end{aligned} \quad (1)$$

This says an action *left*($z$) moves the robot $z$ units to the left (towards the wall) but that the motion stops if the robot hits the wall. It is also assumed that *left*($z$) is the only action that affects $h$. Of course, to move away from the wall, $z$ can be any negative value. Similarly, for $v$, we have:

$$\begin{aligned} v(do(a, s)) = u \equiv \\ \neg \exists z(a = \mathit{up}(z)) \wedge u = v(s) \vee \\ \exists z(a = \mathit{up}(z) \wedge u = v(s) + z). \end{aligned} \quad (2)$$

This captures the upward motion of the robot, while assuming that *up*($z$) is the only action affecting $v$.

Finally, we specify the likelihood axioms in $\mathcal{D}$. We will suppose that the sonar unit, which senses $h$, is quite accurate:

$$\begin{aligned} l(\mathit{sonar}(z), s) = u \equiv \\ (z \geq 0 \wedge u = \mathcal{N}(h(s) - z; 0, .25)) \\ \vee (z < 0 \wedge u = 0) \end{aligned} \quad (3)$$

which stipulates that the difference between a nonnegative reading of $z$ and the true value $h$ is normally distributed with a variance of .25 and mean of 0. (A mean of 0 indicates that there is no systematic bias in the reading.) For the GPS device, assuming that its absolute readings of latitude and longitude have been converted to relative readings (Hightower and Borriello 2001) for $h$ and $v$, imagine a bivariate Gaussian error model:

$$l(\mathit{gps}(x, y), s) = \begin{cases} \mathcal{N}(h(s) - z, v(s) - y; \mu_1, \Sigma) & \text{if } h(s) \geq 2 \\ \mathcal{N}(h(s) - z, v(s) - y; \mu_2, \Sigma) & \text{otherwise} \end{cases}$$

where $\Sigma$ is the $2 \times 2$ identity matrix, $\mu_1 = [0 \ 0]^T$ and $\mu_2 = [0 \ 2]^T$. This says that the components of the Gaussian are independent, and that there is systematic bias in the reading for $v$ when the robot is close to the wall (due to a signal obstructions).

---
[8]Initial beliefs can also be specified for $\mathcal{D}_0$ using *Bel* directly.

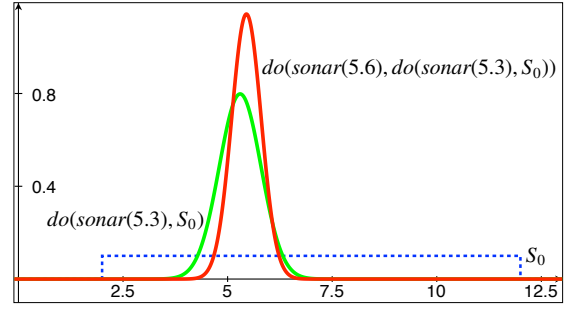

Figure 2: Belief density change for $h$ at $S_0$ (in blue), after sensing 5.3 (in green), and after finally reading 5.6 (in red).

As mentioned earlier, physical actions such as *left*($z$) and *up*($z$) are assumed to be deterministic for this paper, so they are given trivial likelihoods:

$$l(\mathit{left}(z), s) = 1,$$
$$l(\mathit{up}(z), s) = 1.$$

This completes the specification of $\mathcal{D}$.

**Theorem 1:** *The following are logical entailments of $\mathcal{D}$:*

**Initial beliefs**

1. $\mathit{Bel}(\mathit{true}, S_0) = 1$.

2. $\mathit{Bel}(h = 2 \vee h = 3 \vee h = 4, S_0) = 0$
   Although we are integrating a density function $q(x_1, x_2)$ over all real values, $q(x_1, x_2) = 0$ unless $x_1 \in \{2, 3, 4\}$.

3. $\mathit{Bel}(5 \leq h \leq 5.5, S_0) = .05$
   We are integrating a function that is 0 except when $5 \leq x_1 \leq 5.5$. This is $\int_{\mathbb{R}} \int_5^{5.5} .1 \times \mathcal{N}(x_2; 0, 16) \, dx_1 \, dx_2 = .05$.

**Sensing by sonar**

4. $\mathit{Bel}(5 \leq h \leq 5.5, do(\mathit{sonar}(5.3), S_0)) \approx .38$
   Compared to item 3, belief is sharpened significantly by obtaining a reading of 5.3 on the highly sensitive sonar. This is because the $p$ function incorporates the likelihood of a *sonar*(5.3) action. Starting with the density function in item 3, the sensor reading multiplies the expression to be integrated by $\mathcal{N}(x_1 - 5.3; 0, .25)$, as given by (3). This amounts to evaluating the expression

$$\int_{\mathbb{R}} \int_A .1 \times \mathcal{N}(x_1 - 5.3; 0, .25) \times \mathcal{N}(x_2; 0, 16) \, dx_1 \, dx_2$$

with $A = [5, 5.5]$ for the numerator, and $A = [2, 12]$ for the denominator.

5. $\mathit{Bel}(4.5 \leq h \leq 6.5, do[\mathit{sonar}(5.3), \mathit{sonar}(5.6)], S_0) \approx .99$
   Two successive readings around 5.5 sharpen belief within 1 unit of 5.5 to almost certainty. Compared to item 4, the density function is further multiplied by $\mathcal{N}(x_1 -$

5.6; 0, .25), and integrated over [2, 12] for the denominator as usual but over [4.5, 6.5] for the numerator. These changing densities are shown in Figure 2.

**Physical actions**

6. $Bel(h = 0, do(left(4), S_0)) = .2$

   Here a *continuous* distribution evolves into a *mixed* one. By (1), $h = 0$ holds after the action iff $h \leq 4$ held before. This results in $\int_{\mathbb{R}} \int_2^4 .1 \times \mathcal{N}(x_2; 0, 16)\, dx_1\, dx_2 = .2$.

7. $Bel(h \leq 5, do(left(4), S_0)) = .7$

   *Bel*'s definition is amenable to a set of $h$ values, where one value has a weight of .2, and all the other real values have a uniformly distributed density of .1. This change in weights is shown in Figure 3.

8. $Bel(h = 4, do([left(4), left(-4)], S_0)) = .2$
   $Bel(h = 4, do([left(-4), left(4)], S_0)) = 0$

   The point $h = 4$ has 0 weight initially (like in item 2). Moving leftwards *first* means many points "collapse", and so this point (now having $h$ value 0) gets .2 weight which is retained on moving away. But not vice versa.

9. $Bel(-1 \leq v \leq 1, do(left(6), S_0)) =$
   $Bel(-1 \leq v \leq 1, S_0) = \int_{-1}^1 \mathcal{N}(x_2; 0, 16) dx_2 \approx .19$

   Owing to Reiter's solution to the frame problem, belief in $v$ is unaffected by a lateral motion. For $v \in [-1, 1]$ it is the area between $[-1, 1]$ bounded by the specified Gaussian.

10. $Bel(v \leq 1.5, do(up(3.5), S_0)) = Bel(v \leq -2, S_0)$

    After the action $up(3.5)$, the Gaussian for $v$'s value has its mean "shifted" by 3.5 because the density associated with $v = x_2$ initially is now associated with $v = x_2 + 3.5$.

**Sensing by GPS**

11. $Bel(-1 \leq v \leq 1, do(gps(5, .1), S_0)) \approx .27$

    Compared to item 9, a GPS reading of .1 increases the posterior belief for $v \in [-1, 1]$ to $\approx .27$. Using the error model, this is a result of

    $$\int_2^{12} \int_A .1 \cdot \mathcal{N}(x_2; 0, 16) \cdot \mathcal{N}(x_1 - 5, x_2 - .1; \mu_1, \Sigma)\, dx_2\, dx_1$$

    with $A = [-1, 1]$ for the numerator and $A = [-\infty, \infty]$ for the denominator.[9]

**Competing sensors**

12. $Bel(5 \leq h \leq 5.5, do([gps(5, .1), gps(5.3, .1)], S_0)) \approx .27$
    $Bel(5 \leq h \leq 5.5, do([sonar(5.3), gps(5, .1)], S_0)) \approx .42$

    The sonar is more sensitive than the GPS, and so its reading is far more effective. Relating this to item 4, a GPS reading of 5 for $h$ only slightly redistributes the density.

**Systematic bias**

13. $h(S_0) \leq 4 \supset$
    $Bel(-1 \leq v \leq 1, do([left(4), gps(1, 0)], S_0)) \approx 0$

---

[9]This is a simple instance of Kalman filtering (Dean and Wellman 1991) where the value being sensed is static. Gaussian distributions enjoy the conjugate property: multiplying Gaussians results in another Gaussian (Box and Tiao 1973), and is easily computed.

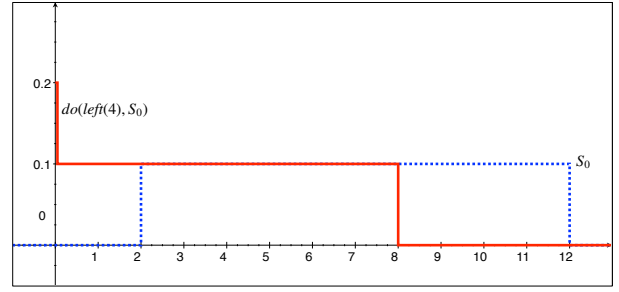

Figure 3: Belief update for $h$ after physical actions. Initial belief at $S_0$ (in blue) and after a leftward move by 4 (in red).

After moving left by 4 units, $v$'s reading from the GPS has a systematic bias of 2. Among other things, this entails that the belief in $v \leq 1$ is almost 0 which is much weaker than its prior from item 9.

**Nonstandard properties**

14. $Bel(h > 7v, S_0) \approx .6$

    Beliefs about any mathematical expression involving the random variables, even when that does not correspond to well known density functions, are entailed. In this case, we are basically evaluating:

    $$\int_2^{12} \int_{-\infty}^{x_1/7} .1 \times \mathcal{N}(x_2; 0, 16)\, dx_2\, dx_1.$$

15. $Bel([\exists a, s.\ now = do(a, s) \wedge h(s) > 1], do(left(4), S_0)) = 1$.

    It is possible to refer to earlier or later situations using *now* as the current situation. This says that after moving, there is full belief that ($h > 1$) held before the action.

## Related Work

Sensor fusion has been a primary concern in state estimation approaches (Thrun, Burgard, and Fox 2005). Popular models include variants of Kalman filtering (Fox et al. 2003), where priors and likelihoods are assumed to be Gaussian. We already pointed out that entailment item 11 is a simple instance of Kalman filtering. But in general, our approach does not make any assumptions about the nature of distributions, nor about how distributions and dependencies may evolve after actions, and allows for strict uncertainty. This distinguishes the current method from numerous probabilistic formalisms (Lerner et al. 2002; Dean and Wellman 1991; Fox et al. 2003), including those that handle explicit actions (Darwiche and Goldszmidt 1994; Hajishirzi and Amir 2010). To the best of our knowledge, none of these formalisms have treated cases where state variables change in the manner indicated in the paper.

In the rest of the section, we will briefly discuss how the framework used in the paper is related to existing logical formalisms for uncertainty. Probabilistic logical formalisms such as (Halpern 1990; Bacchus 1990) are equipped to handle disjunctions and quantifiers, but they do not explicitly address actions. Relational probabilistic languages and Markov logics (Ng and Subrahmanian 1992; Richardson and Domingos 2006) also do not model actions. (Recent temporal extensions, such as (Choi, Guzman-Rivera, and Amir 2011), specifically treat Kalman filtering, and not complex actions.) In this regard, action logics such as dynamic and process logics are closely related. Recent proposals, for example (Van Benthem, Gerbrandy, and Kooi 2009), treat sensor fusion. However, these and related frameworks (Halpern and Tuttle 1993), including probabilistic planning formalisms (Kushmerick, Hanks, and Weld 1995), are mostly propositional. Proposals based on the situation and fluent calculi are first-order (Bacchus, Halpern, and Levesque 1999; Poole 1998; Boutilier et al. 2000; Mateus et al. 2001; Shapiro 2005; Gabaldon and Lakemeyer 2007; Fritz and McIlraith 2009; Belle and Lakemeyer 2011; Thielscher 2001), but none of them deal with continuous sensor noise, and nor do the extensions for continuous processes (Reiter 2001; Herrmann and Thielscher 1996). Moreover, none of these deal with the integration of continuous variables within the language.

## Conclusions

This paper illustrates location estimation for a robot operating in an incompletely known world, equipped with noisy sensors. In contrast to a number of competing formalisms, where the modeler is left with the difficult task of deciding how the dependencies and distributions of state variables might evolve, here one need only specify the initial beliefs and the physical laws. Suitable posteriors are then entailed. The framework of the situation calculus, and a recent generalization to the BHL scheme, allows us to additionally specify situation-specific biases and realistic continuous error models. Our example demonstrates that belief changes appropriately even when one is interested in nonstandard properties, such as logical relationships of state variables, all of which emerges as a side-effect of the general specification. In the future, we intend to consider the more elaborate case where a robot's position will include angular orientation in addition to Cartesian coordinates, and explore state estimation in this setting. On the more computational side, we are interested in investigating formal conditions about action theories that would allow us to estimate posteriors efficiently under the assumption that priors and likelihoods are drawn from tractable distributions (Box and Tiao 1973).

## Acknowledgements

The authors would like to thank the referees for constructive feedback, and the Natural Sciences and Engineering Research Council of Canada for financial support.

## Appendix: Sums and Integrals in Logic

Logical formulas can be used to characterize sums and a variety of sorts of integrals. Here we show the simplest possible cases: the summing of a one variable function from 1 to $n$, and the definite integral from $-\infty$ to $\infty$ of a continuous real-valued function of one variable. Other complications are treated in a longer version of the paper.

First, sums. For any logical term $t$ and variable $i$, we introduce the following notation to characterize summations:

$$\sum_{i=1}^{n} t = z \doteq \exists f \ [f(1) = t_1^i \wedge f(n) = z \wedge \\ \forall j \ (1 \leq j < n \supset f(j+1) = f(j) + t_{(j+1)}^i )]$$

where $f$ is assumed to not appear in $t$, and $j$ is understood to be chosen not to conflict with any of the variables in $t$ and $i$.

Now, integrals. We begin by introducing a notation for limits to positive infinity. For any logical term $t$ and variable $x$, we let $\lim_{x \to \infty} t$ stand for a term characterized by:

$$\lim_{x \to \infty} t = z \doteq \forall u (u > 0 \supset \exists m \ \forall n (n > m \supset \left| z - t_n^x \right| < u)).$$

The variables $u$, $m$, and $n$ are understood to be chosen here not to conflict with any of the variables in $x$, $t$, and $z$.

Then, for any variable $x$ and terms $a$, $b$, and $t$, we introduce a term INT$[x, a, b, t]$ to stand for the definite integral of $t$ over $x$ from $a$ to $b$:

$$\text{INT}[x, a, b, t] \doteq \lim_{n \to \infty} h \cdot \sum_{i=1}^{n} t_{(a+h \cdot i)}^x$$

where $h$ stands for $(b - a)/n$. The variable $n$ is chosen not to conflict with any of the other variables.

Finally, we define the definite integral of $t$ over all real values of $x$ by the following:

$$\int_x t \doteq \lim_{u \to \infty} \lim_{v \to \infty} \text{INT}[x, -u, v, t].$$

The main result for this logical abbreviation is the following:

**Theorem 2:** *Let $g$ be a function symbol of $\mathcal{L}$ standing for a function from $\mathbb{R}$ to $\mathbb{R}$, and let $c$ be a constant symbol of $\mathcal{L}$. Let $M$ be any logical interpretation of $\mathcal{L}$ such that the function $g^M$ is continuous everywhere. Then we have the following:*

$$\text{If} \int_{-\infty}^{\infty} g^M(x) \, . \, dx = c^M \ \ \text{then} \ \ M \models (c = \int_x g(x)).$$